\documentclass{article}

\usepackage{arxiv}

\usepackage[utf8]{inputenc} 
\usepackage[T1]{fontenc}    
\usepackage{hyperref}       
\usepackage{url}            
\usepackage{booktabs}       
\usepackage{amsfonts}       
\usepackage{amssymb}
\usepackage{graphicx}
\usepackage{algorithm}
\usepackage{algorithmic}
\usepackage{lipsum}		

\title{Wasserstein Distance Based Domain Adaptation for Object Detection}

\date{} 					

\author{
  Pengcheng Xu \\
  University of Maryland College Park\\
  College Park, MD 20742 \\
  \texttt{alexxu@umd.edu} \\
   \And
 Prudhvi Gurram \\
  Booz Allen Hamilton\\
  CCDC Army Research Laboratory\\
  Adelphi, MD 20783 \\
  \texttt{gurram\_prudhvi@bah.com} \\
   \And
 Gene Whipps \\
  CCDC Army Research Laboratory\\
  Adelphi, MD 20783 \\
  \texttt{gene.t.whipps.civ@mail.mil} \\
   \And
 Rama Chellappa \\
  University of Maryland College Park\\
  College Park, MD 20742 \\
  \texttt{rama@umiacs.umd.edu} \\
}

\renewcommand{\headeright}{}
\renewcommand{\undertitle}{}

\begin{document}
\maketitle

\begin{abstract}
In this paper, we present an adversarial unsupervised domain adaptation framework for object detection. Prior approaches utilize adversarial training based on cross entropy between the source and target domain distributions to learn a shared feature mapping that minimizes the domain gap. Here, we minimize the Wasserstein distance between the two distributions instead of cross entropy or Jensen-Shannon divergence to improve the stability of domain adaptation in high-dimensional feature spaces that are inherent to object detection task. Additionally, we remove the exact consistency constraint of the shared feature mapping between the source and target domains, so that the target feature mapping can be optimized independently, which is necessary in the case of significant domain gap. We empirically show that the proposed framework can mitigate domain shift in different scenarios, and provide improved target domain object detection performance.
\end{abstract}

\keywords{Unsupervised domain adaptation \and Object detection \and Wasserstein distance}

\section{Introduction}
\label{sec:intro}
Object detection is a fundamental computer vision task that aids in scene understanding using visual sensors. It involves locating and classifying all instances of different objects such as people, vehicles, etc. in images. With the advent of large datasets, e.g. ImageNet \cite{imagenet}, there is a need for algorithms that can learn the basic concepts that make up a particular object by learning diverse feature representations of different objects. This led to the rise of deep neural networks (DNN) based object detection algorithms \cite{girshick2014,girshick2015,ren2015,he2015,he2017,redmon2016,liu2016} that give a significantly improved detection performance over the previously developed algorithms that use hand-crafted features \cite{dalal2005histograms}. The DNN-based algorithms are able to learn the basic concepts or attributes of different objects by encoding the hidden semantic information among the pixels in the millions of parameters in a single DNN architecture \cite{zeiler2014visualizing}. There are two categories of DNN-based object detection algorithms: one-step and two-step. In two-step algorithms such as Faster-RCNN, Mask-RCNN, and FPN \cite{girshick2014,girshick2015,ren2015,he2015,he2017}, the detector extracts feature maps of images in the first step, and generates coarse region proposals based on the feature maps, and then classifies each proposal into different objects or background in the second step. In one-step algorithms such as YOLO \cite{redmon2016} and SSD \cite{liu2016}, the detector performs feature extraction, bounding box regression and object classification together in a single step, thereby significantly reducing computational time cost. So, one-step algorithms are more suitable for real-time object detection.  

During the training process, the back-bone neural network of a detector (feature generator) is first pre-trained on an image classification dataset such as ImageNet \cite{imagenet}, and then, the entire network is trained using object detection datasets such as Pascal-VOC \cite{pascal-voc-2012} or MS-COCO \cite{mscoco} that have images with annotated bounding boxes and labels of objects. For good generalization performance, the DNN-based object detection algorithms need a large number of such annotated images to train the millions of parameters. Even in this case, the detectors will give good performance on such data that exhibits similar feature representations and distributions as the training data. The performance of the algorithms will degrade when they are applied on data that is significantly different from the training data. Henceforth, we will call the large datasets, on which object detectors are trained as the source domain, and the data collected in test conditions as the target domain. Some examples of the different train-test conditions are different weather conditions, different sensors, real and synthetic imagery, etc. 

In some cases, the DNN-based algorithms can be "fine-tuned" by using several hundred or thousands of labeled samples from the target domain. However, manually locating and labeling objects in images requires large amount of time and labor, which may not be realistic in practical applications. There are also weakly-supervised and semi-supervised algorithms \cite{tang2016,bilen2016,inoue2018}, in which a small number of annotations or label information from the target domain are available. However, more often than not, it is not possible to obtain labeled data from the target domain. Hence, in this paper, we propose to perform unsupervised domain adaptation that transfers the knowledge learned by the DNN-based object detector from the source domain to the target domain so that the performance of the detector in target domain is improved without the need for labeled training data from the target domain.

\section{Unsupervised Domain Adaptation}
\label{sec:motivation}
Unsupervised domain adaptation (UDA) bridges the domain shift between the source and target domains, where only the source label information is available. Until recently, research in UDA has been focused on the applications of image classification and semantic segmentation. Our work focuses on UDA for object detection. Several algorithms \cite{ganin2014,tzeng2017,tzeng2014,long2015,saito2018cvpr,balaji2019} have been proposed to mitigate the domain shift by minimizing different kinds of distance measures between the source and target feature distributions. Inspired by the generative adversarial networks (GAN) \cite{goodfellow2014}, recently adversarial domain adaptation algorithms have gained popularity. In such algorithms, the feature extractor is considered as the generator, and the source/target domain classifier is considered as the discriminator. The feature extractors are trained to generate similar distributions for both source and target domains to fool the domain classifier. The domain classifier is trained to distinguish the feature distributions from the source and target domains. After the training is completed, the feature extractor should be able to generate feature distributions for source and target domains in such a way that the domain classifier cannot distinguish them, thus achieving domain adaptation. They are considered adversarial because the feature extractor and the domain classifier are treated as adversaries to each other during the training process. 

In one such technique, Ganin and Lempitsky \cite{ganin2014} inserted a gradient reversal layer (GRL) between the feature extractor and the domain classifier to flip the sign of the gradient from the domain classifier before back propagating to the feature extractor and performing adversarial training by pushing the feature extractor and the domain classifier in opposite directions. This algorithm was used to implement domain adaptation for image classification task in \cite{ganin2014}. Following the same idea, two recent approaches applied GRL-based domain adaptation for object detection task \cite{chen2018,saito2018cvpr} using Faster-RCNN framework. In both of these, two domain classifiers were used to align feature distributions at two different image levels. However, they hold different opinions on the validity of using the features generated by the region proposal network (RPN) for domain alignment. In \cite{chen2018}, the authors applied an instance-level alignment to the region-of-interest (ROI) features aiming to change the RPN for domain adaptation. In \cite{saito2018cvpr}, the authors argue that the proposals generated by the source network may not localize objects in the target domain accurately, which may degrade the adaptation performance. However, GRL requires the source and target domain feature extractors to share the same neural network which is not necessarily an amiable constraint for domain adaptation. This constraint reduces the number of parameters of the framework but enforces the exact consistency between the source and target mappings, which makes the optimization poorly conditioned \cite{tzeng2017}. Instead of GRL, we follow the adversarial training scheme of GAN, but do not enforce any consistency constraint between the source and target feature extractors.

In adversarial domain adaptation, the metric that is used to measure the distance between the source and target domain feature distributions, is crucial for the performance of the domain discriminator and the final adapted model. The adversarial discriminative domain adaptation \cite{tzeng2017} uses a domain classifier with cross entropy (CE) loss to distinguish the feature vectors from the source domain and the target domain. The minimization of CE between two distributions for domain adaptation is equivalent to the minimization of the Kullback-Leibler (KL) divergence between the two distributions \cite{arjovsky2017}. When the feature distributions are low-dimensional manifolds in a high-dimensional space, it is likely that the distributions might have a negligible intersection, and Kullback-Leibler (KL) divergence is not defined. This is of great concern in object detection models, where the ambient feature space of the backbone network usually has very high dimensionality. To avoid this, one can use Jensen-Shannon (JS) divergence, which is a symmetrized and smoothed version of KL divergence, that measures the sum of KL divergences between the respective distributions and the mean of the two distributions. However, the discriminator may learn very quickly to distinguish between the two domain distributions, and the JS divergence may be locally saturated. This may lead to the vanishing gradient problem \cite{arjovsky2017}. CE, KL and JS divergence are not very stable metrics for domain adaptation in high-dimensional spaces. So, we use Wasserstein distance \cite{arjovsky2017,balaji2019} as the distance metric between the source and target distributions in our discriminator. The Wasserstein distance or Earth-Mover distance between two distributions, $P_s$ and $P_t$ is given by
\begin{equation}
W\left(P_s, P_t\right) = \inf_{\gamma\in\Pi\left(P_s, P_t\right)} E_{(x,y) \thicksim \gamma} \left[\left\|x - y\right\|\right].
\end{equation}
However, as stated in \cite{arjovsky2017}, this infimum (greatest lower bound) is intractable and so, one can use the Kantorovich-Rubinstein duality of the Wasserstein distance, which is given by
\begin{equation}
\label{eq:wd}
W\left(P_s, P_t\right) = \sup_{\|f\|_L \leq 1} E_{x \thicksim P_s} \left[f\left(x\right)\right] - E_{x \thicksim P_t} \left[f\left(x\right)\right],
\end{equation}
where the supremum (least upper bound) is over all 1-Lipschitz functions $f:X \rightarrow \mathbb{R}$. If $f$ is a K-Lipschitz function, then the above equation leads to a scaled version of Wasserstein distance $ KW\left(P_s, P_t\right)$, which is also a valid metric to optimize for domain adaptation. We detail how we use this metric for domain adaptation in Section~\ref{sec:ga}. 

In summary, we propose an unsupervised adversarial domain adaptation framework for Faster-RCNN. We chose Faster-RCNN as the base object detection framework because of its superior performance, and the ease of extensibility. We align the source and target domains on both the global feature level and the region proposal feature (local) level using the Wasserstein distance metric. For the global feature level alignment, the source and target networks do not share any layers for the backbone feature generator. For the local feature level alignment, we incorporate the object detection loss to stabilize the adversarial training similar to \cite{odena2017,sankaranarayanan2018learning}. Hence, the source and target domain networks share the same RPN and the classifier.

\section{Proposed Method}
\label{sec:propmeth}
The overview of our proposed domain adaptive Faster-RCNN framework is shown in Fig.1. In Phase 1, we extract the global feature maps of imagery from source and target domains using their individual backbone feature generator networks, and perform the alignment of the feature spaces using the Wasserstein distance measure. Then, in Phase 2, the aligned global feature maps are fed into the RPN to generate region proposals and extract ROI feature maps. We perform the local alignment of ROI proposal features between the source domain and target domain, again using the Wasserstein distance measure. At the same time, the object detection loss from the labeled source domain is used to stabilize and introduce categorical information into the adversarial domain adaptation training. Since the Faster-RCNN is a two-step algorithm, we denote the backbone feature generator networks of the source and target domain as global mappings $M_s^g$ and $M_t^g$, the shared RPN as local mapping $M^l$, and the shared classifier as $C$. Here, we assume that we have the source images $X_s$ and the corresponding labels or annotations $Y_s$ drawn from a distribution $p_s(x,y)$. The target domain images $X_t$ and labels $Y_t$ are drawn from a distribution $p_t(x,y)$. However, for the target domain, we only have the images $X_t$, but not the labels $Y_t$. The objective is to learn a target domain mapping $M^g_t$ that can map the target domain images $X_t$ to global feature space, and adapt the RPN $M^l$ and the classifier $C$, to generate region proposals and assign right labels $Y_t$, thus generating object annotations for the target images $X_t$, using just the knowledge of source domain mappings $M^g_s$, $M^l$ and classifier $C$. 

\begin{figure*}
\begin{center}
\includegraphics[width=\textwidth]{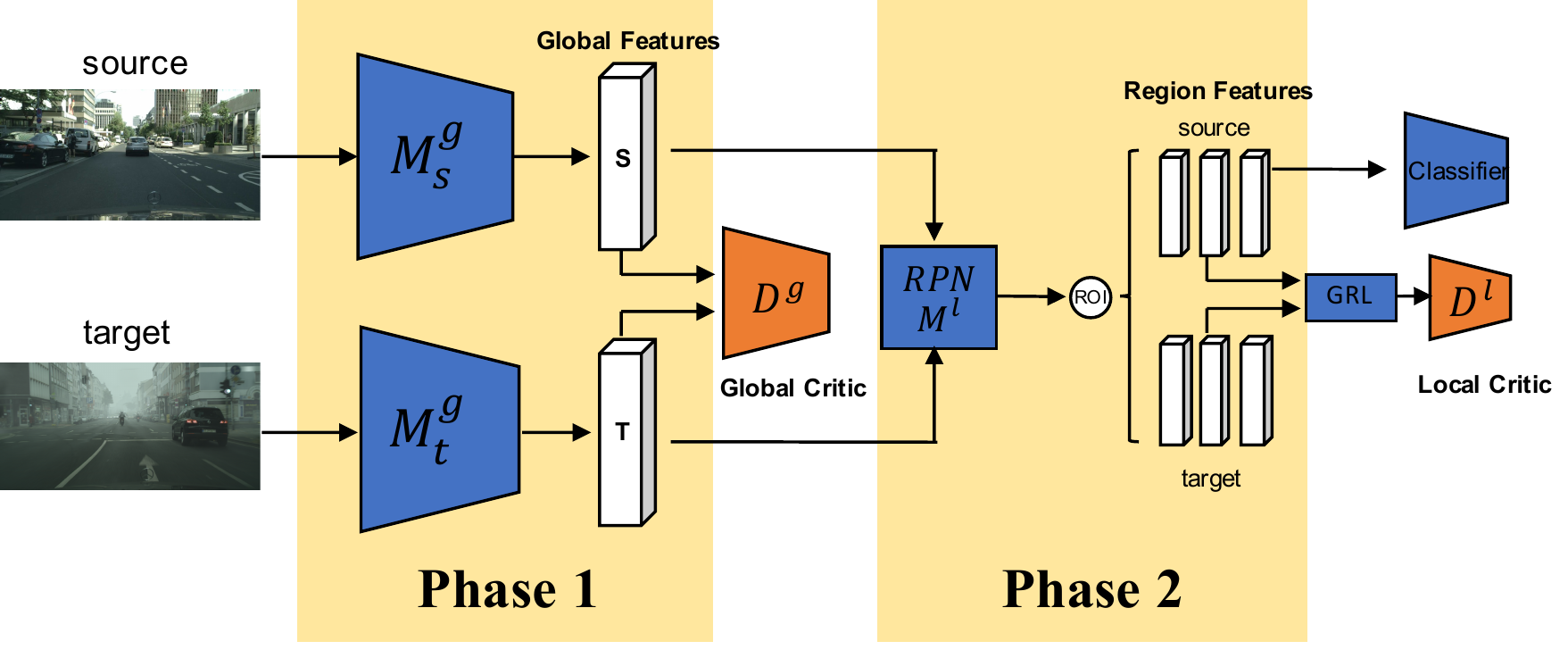}
\end{center}
\caption{Proposed framework.}
\label{fig:framework}
\end{figure*}

\subsection{Phase 1 - Global Alignment}
\label{sec:ga}
In Phase 1, first we pass the images from the source domain $X_s$ and from the target domain $X_t$ through the backbone feature extractor networks represented by $M_s^g$ and $M_t^g$ respectively to obtain their global feature representations $M_s^g(X_s)$ and $M_t^g(X_t)$. The goal of global alignment is to minimize the distance between the distributions of the source and target global features. The source domain object detector is pre-trained on the source domain imagery and the source global mapping $M_s^g$ is fixed during global alignment. As mentioned in Section~\ref{sec:motivation}, unlike \cite{chen2018} and \cite{saito2018cvpr}, the feature generators of the source and target networks, $M_s^g$ and $M_t^g$ do not share the parameters in any layer, which intuitively gives more freedom for the target global mapping to explore the parameter landscape and find the optimal parameters that would push $M_t^g(X_t)$ as close as possible to $M_s^g(X_s)$. We initialize the target global mapping $M_t^g$ with that of the fixed source mapping $M_s^g$. 

Next, we estimate the Wasserstein distance between the source domain and target domain distributions $M_s^g(X_s)$ and $M_t^g(X_t)$ using Eq.~\ref{eq:wd}. We use another neural network $D^g$, similar to the discriminator in GAN, to encode the 1-Lipschitz continuous functions $f$. So the Wasserstein distance between the two distributions can be written as
\begin{equation}
W(M_s^g,M_t^g) = \sup_{\|D^g\|_L \leq 1} \mathbb{E}_{X_s \thicksim p_s}(D^g(M_s^g(X_s))) - \mathbb{E}_{X_t \thicksim p_t}(D^g(M_t^g(X_t))).
\end{equation}
We can optimize for the parameters of the network $D^g$, and thus the optimal 1-Lipschitz function, by back-propagating the gradient of the right-hand side term in the above equation. The encoding $D^g$ is called as \textit{critic} in the context of Wasserstein distance. In \cite{arjovsky2017}, the parameters of $D^g$ are clipped after every update to satisfy the Lipschitz continuity requirement. Instead, here we use spectral normalization \cite{miyato2018} for each layer in the global critic network to enforce this requirement. Also, to mitigate the Wasserstein distance estimation problem that arises due to small training batch size used in object detection, we use a patch-based critic network similar to those used in \cite{chen2018} and \cite{johnson2016} to estimate the Wasserstein distance. In particular, we treat each activation of the last convolution layer of the critic network as a feature map of one patch of an image. 

Finally, after the Wasserstein distance is estimated, we can minimize this distance measure by optimizing for the parameters of the target global mapping or feature extractor $M_t^g$. The two-step iterative optimization for global alignment can be formulated as following:
\begin{equation}
\max_{D^g} W(M_s^g,M_t^g) = \frac{1}{n} \sum_{i} D^g(M_s^g(X_s^i)) - \frac{1}{n} \sum_{i} D^g(M_t^g(X_t^i)),
\end{equation}
\begin{equation}
\min_{M_t^g} W(M_s^g,M_t^g) = - \frac{1}{n} \sum_{i} D^g(M_t^g(X_t^i)),
\end{equation}
where $n$ is the batch size of one mini-batch. Since the source global mapping $M_s^g$ is fixed, we update only the target global mapping $M_t^g$ in the second step.

\subsection{Phase 2 - Local Alignment}
\label{sec:la}
In global alignment, we optimize the backbone feature extractor of target network to align the global feature distributions of the source and target domains. In Phase 2, we further align the distributions of local ROI features in the two domains. We also would like to use the source domain object detection loss to stabilize the adversarial training process. So, for the local alignment, we constrain the source and target networks to share the same RPN and classifier as shown in Fig.~\ref{fig:framework}. We use Wasserstein distance to measure the domain gap at the local level as well. So, we use a neural network $D^l$, similar to $D^g$ in global alignment step, to encode the \textit{critic} that estimates the distance between the source and target domain distributions at the local level. The traditional adversarial technique that we used to update the feature generator at the global level cannot be used here as the source and target networks share the RPN. So, we use GRL \cite{ganin2014} to update the local critic network and the RPN (feature generator) at the same time. Here, we denote global features of an image $i$ from the backbone networks of source and target networks as $f_s^i$ and $f_t^i$. Then loss functions for local alignment can be formulated as following:
\begin{equation}
\max_{D^l} W(M^l(f_s),M^l(f_t)) = \frac{1}{nm} \sum_{i,j} D^l(M_{j}^{l}(f_s^i)) - \frac{1}{nm} \sum_{i,j} D^l(M_{j}^l(f_t^i)),
\end{equation}
\begin{equation}
\min_{M^l} W((M^l(f_s),M^l(f_t)) = \frac{1}{nm} \sum_{i,j} D^l(M_{j}^{l}(f_s^i)) - \frac{1}{nm} \sum_{i,j} D^l(M_{j}^l(f_t^i)),
\end{equation}
where $n$ is the batch size (number of images) of one mini-batch and $m$ is the number of proposals generated by the RPN for each image. One can observe that the loss functions of the local critic network $D^l$ and the RPN (feature generator) are exact opposite of each other. In GRL, the gradient that is used to update the local discriminator network is reversed in sign, and then used to update the RPN. We also iteratively update the classifier along with the RPN to minimize the source domain object detection loss $L_{det}$ while performing the local alignment. 

%
%
%

\section{Experimental Results}
\label{sec:results}
In this section, we evaluate the adaptation performance of the proposed framework on two domain-shift scenarios: different weather conditions (Cityscapes \cite{cordts2016} to Foggy Cityscapes \cite{sakaridis2018}), and between synthetic and real imagery (SIM10k \cite{johnson2016} to Cityscapes). We compare the performance of our proposed framework with that of four algorithms: Faster-RCNN, domain adaptive Faster-RCNN (DA-Faster) \cite{chen2018}, BDC-Faster and the strong and weak alignment Faster-RCNN (SW-Faster) \cite{saito2018cvpr}. The Faster-RCNN is the model trained on the source data and evaluated on the target data without any domain adaptation. BDC-Faster is the Faster-RCNN model with only a global domain classifier that is optimized using CE loss, which was originally reported in \cite{saito2018cvpr}. We included this particular model in our experiments because the comparison between our model and BDC-Faster shows the advantages of using Wasserstein distance over the CE loss for domain adaptation. Since our task focuses on unsupervised domain adaptation in feature space, for a fair comparison, we do not compare our model to the one of the models proposed in \cite{saito2018cvpr}, where additional images are generated by using cycle GAN \cite{CycleGAN2017}, and the additional images are then used for domain adaptation. For all the experiments, we use the Faster-RCNN as the base detector and the VGG16 \cite{simonyan2014} as the backbone feature generator network. The algorithm, implementation details including the global and local critic network architectures are presented in the supplementary material.

\subsection{Adaptation across different weather conditions}
\label{sec:city2foggy}
In this experiment, we evaluate the adaptation ability of our model under different weather conditions. We use the Cityscapes as the source domain and the Foggy Cityscapes as the target domain. The Cityscapes dataset includes the images of street scenes of European cities captured in clear weather. The original size of each image is 2048 $\times$ 1024. The whole dataset includes 2,975 training images and 500 test images which cover 8 object categories: \textit{person, rider, car, truck, bus, train, motorcycle,} and \textit{bicycle}. The Foggy Cityscapes dataset is rendered from Cityscapes by adding the fog noise. The images in the target domain are generated from the source domain, which implies that the source and target domains share the same semantic information. Since the original Cityscapes dataset was created for semantic segmentation, and does not have any bounding-box annotation, similar to \cite{chen2018}, we manually create the ground truth bounding-box annotations for each object by taking the tightest rectangle of its instance mask.

\textbf{Results:} As shown in Table~\ref{tab:city2foggy}, the proposed model with just global alignment can achieve an mAP of 32.0\%, which is much better than the performance of the baseline model without any adaptation. It is also better than the BDC-Faster which uses a global domain classifier with the CE loss. This reinforces our argument that the Wasserstein distance is not only more stable but also more effective compared to the CE loss for domain adaptation. Our model is also better than the DA-Faster and the SW-Faster with only global alignment and regularization term. This confirms our argument that the source and target feature extractor networks sharing the same parameters may limit the adaptation performance. From the results, we can also see that local alignment does not contribute as much as the global alignment to the final performance of the model. This is expected in this experiment because Cityscapes and Foggy Cityscapes share exactly the same bounding-box annotations and categorical information. So, conditioned on the aligned global feature distributions, the distribution of bounding-box annotations and labels for these two datasets would be very similar. The local alignment, which is meant to align the distributions of bounding box locations and labels between the two domains would not be very helpful.

\begingroup
\begin{table}
\begin{center}
\setlength{\tabcolsep}{2.5pt}
\begin{tabular}{|l|cccc|ccccccccc|}
\hline
Method & G &I &L &C &bus &bicycle &car &mcycle &person &rider &train &truck &mAP\\
\hline\hline
Faster-RCNN & & & & &24.2 &26.4 &31.6 &14.0 &24.0 &32.0 &9.1 &11.7 &21.6\\
BDC-Faster & \checkmark & & & &29.2 &28.9 &42.4 &22.6 &26.4 &37.2 &12.3 &21.2 &27.5\\
DA-Faster & \checkmark &\checkmark & & &25.0 &31.0 &40.5 &22.1 &35.3 &20.2 &20.0 &27.1 &27.6\\
SW-Faster & \checkmark & & &\checkmark &38.0 &31.2 &41.8 &20.7 &26.6 &37.6 &19.7 &20.5 &29.5\\
SW-Faster & \checkmark & &\checkmark &\checkmark &36.2 &35.3 &43.5 &30.0 &29.9 &42.3 &32.6 &24.5 &34.3\\
\hline
Ours & \checkmark & & & &41.6 &32.3 &44.0 &24.4 &30.6 &40.5 &21.5 &21.4 &32.0\\
Ours & \checkmark &\checkmark & & &39.9 &34.4 &44.2 &25.4 &30.2 &42.0 &26.5 &22.2 &33.1\\
\hline
\end{tabular}
\end{center}
\caption{Results on adaptation from Cityscapes to Foggy Cityscapes(\%). G, I, L, C represent the global alignment, local alignment on ROI, local alignment on the intermediate layer of VGG, and context-vector based regularization respectively.}
\label{tab:city2foggy}
\end{table}
\endgroup

\subsection{Adaptation from synthetic to real imagery}
\label{sec:sim2city}
For computer vision tasks such as object detection, manually labeling or annotating real data costs a large amount of time and labor, which is not practical. However, for the synthetic data, it is easy and convenient to generate the ground truth labels and annotations for a large number of objects while generating the imagery. The natural goal here is to transfer the object detection model trained on the synthetic data to the real data. So we use the SIM10k dataset as the source domain and the Cityscapes dataset as the target domain. Following \cite{chen2018}, we evaluate the model only on the AP of \textit{car}. SIM10k is a dataset of the synthetic driving scenes rendered from the game \textit{Grand Theft Auto (GTAV)} which includes 10,000 training images. The model is evaluated on the test set of the Cityscapes (500 images).

\textbf{Results:} As shown in Table~\ref{tab:sim2city}, our model with the global alignment can achieve an AP of 39.8\% which is very close to the state of art result. Moreover, we achieve this result with only a few epochs of global domain alignment training, which is very efficient time-wise. These results confirm that the proposed model can reduce the domain shift between the synthetic data and the real data. In Phase 2, we fix the VGG16 backbone of the target network and perform the local alignment on the RPN. According to the experimental results, the local alignment can further improve the performance since the bounding box distributions of SIM10k and Cityscapes are different.

\begin{table}
\begin{center}
\begin{tabular}{|l|cccc|c|}
\hline
Method & G &I &L &C &AP on car\\
\hline\hline
Faster-RCNN & & & & &34.5 \\
BDC-Faster &\checkmark & & & &31.8 \\
DA-Faster &\checkmark &\checkmark & & &38.9 \\
SW-Faster & \checkmark & & &\checkmark &36.4\\
SW-Faster & \checkmark & &\checkmark &\checkmark &40.1\\
\hline
Ours &\checkmark & & & &39.8\\
Ours &\checkmark &\checkmark & & &40.6\\
\hline
\end{tabular}
\end{center}
\caption{Results on adaptation from SIM10k to Cityscapes(\%)}
\label{tab:sim2city}
\end{table}


%

\section{Discussion}
\label{sec:disc}
\begin{figure}[h]
\begin{center}
\includegraphics[width=\textwidth]{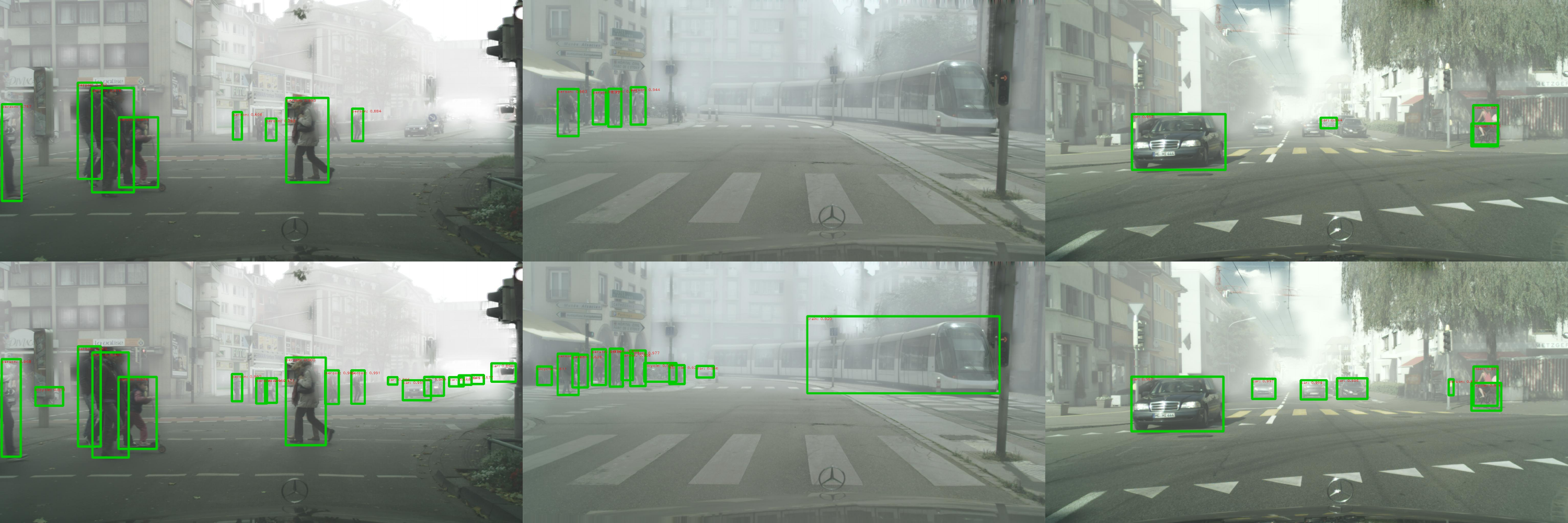}
\end{center}
\caption{Results from Cityscapes to Foggy cityscapes before and after domain adaptation}
\label{fig:sampleimg1}
\end{figure}

\begin{figure}[h]
\begin{center}
\includegraphics[width=\textwidth]{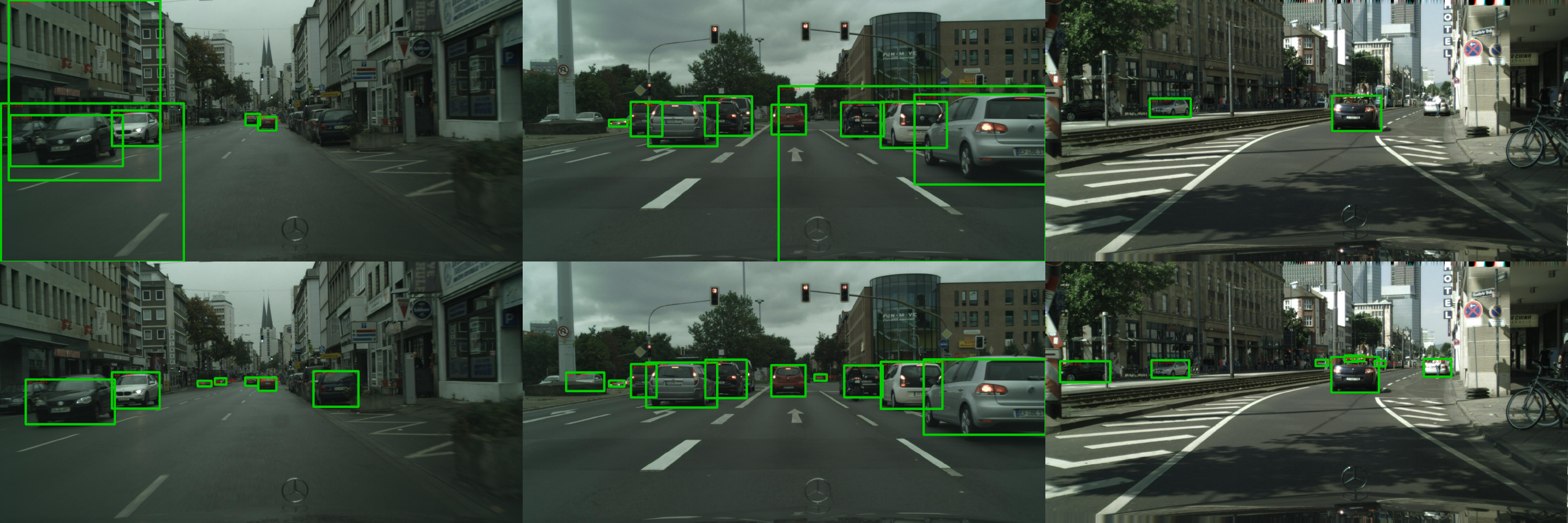}
\end{center}
\caption{Results from Sim10k to Cityscapes before and after domain adaptation}
\label{fig:sampleimg2}
\end{figure}

Figs.~\ref{fig:sampleimg1} and \ref{fig:sampleimg2} illustrate the improvement of object detection results after domain adaptation between Cityscapes-Foggy Cityscapes and between SIM10k-Cityscapes respectively. The images in the top row show the bounding boxes of detected objects before domain adaptation and the corresponding images in the bottom row show the bounding boxes of detected objects after domain adaptation. One can see that the objects that were missed before domain adaptation due to the domain gap, are detected after domain alignment in both cases. The locations of the detected objects are also more accurate after adaptation. Moreover, in the case of synthetic to real imagery (SIM10k-Cityscapes), one can also observe that the number of false detections before domain alignment has been significantly reduced after the adaptation.

From the results presented in Tables~\ref{tab:city2foggy} and \ref{tab:sim2city}, one can see that most of the improvement in object detection performance is obtained after the global alignment step (Phase 1) itself. The reasons for this are the two main contributions of this paper - using Wasserstein distance to measure the domain gap and optimizing for the target feature generator independent of the source network - as explained in Section~\ref{sec:results}. However, the local alignment does not provide significant improvement in the detection performance. One of the reasons is the small batch size used for Wasserstein distance estimation (learning the critic network) due to constraints on the available computational resources (GPU memory). Another reason is that currently, there is no regularization between global and local alignment steps. The local alignment results may be further improved by using a larger batch size for Wasserstein distance estimation, or by simply improving the sample efficiency of Wasserstein distance estimation. Adding a regularization term to unify the training of the global and local domain critic networks may also improve the domain adaptation results.
%

\section{Conclusion}
\label{sec:conc}
We have proposed an unsupervised domain adaptive framework based on Faster-RCNN for object detection. We perform the alignment of the source and target domain distributions at two levels - global and local. We use Wasserstein distance metric to estimate the distance between the distributions of the two domains, which mitigates the gradient vanishing problem in high-dimensional spaces of object detection task. Additionally at the global level, different from previous approaches, the source and target networks in our model do not share any feature extraction layers, which provides more flexibility for the target network to learn the required feature mapping that minimizes domain shift even in the case of significant domain gap. During the local alignment, we also iteratively minimize the object detection loss along with domain adaptation loss to stabilize the adversarial training. We have validated our framework by testing it on datasets exhibiting different domain shift such as different weather conditions, synthetic to real imagery, and comparing it to other unsupervised domain adaptation techniques in terms of object detection performance.

\bibliographystyle{unsrt}  
\bibliography{egbib}  

\newpage
\title{Wasserstein Distance Based Domain Adaptation for Object Detection: Supplementary Material}
\maketitle

\section*{Global and Local Critic Network Architectures}
We summarize the architectures of global and local critic networks in the following tables. For the global critic network, we first reduce the dimensionality of the feature maps by max-pooling, because larger dimensional feature maps require larger capacity of the critic network and larger batch-size of images, which increases the need for larger GPU memory. Due to the limited computational resources, we chose to use smaller feature maps.

\begin{table}[h]
\begin{center}
\begin{tabular}{|c|c|}
\hline
Global critic network\\
\hline
Max Pooling, kernel=2, stride=2, pad=0 \\
Conv1, channel=512x512, kernel=3, stride=2, pad=1 \\
Spectral Normalization, LeakyRelu \\
Conv2, channel=512x128, kernel=3, stride=2, pad=1 \\
Spectral Normalization, LeakyRelu \\
Conv3, channel=128x128, kernel=3, stride=1, pad=1 \\
Spectral Normalization, LeakyRelu \\
Conv4, channel=128x1, kernel=3, stride=1, pad=1 \\
Spectral Normalization\\
\hline
\end{tabular}
\end{center}
\caption{The architecture of the global critic network}
\end{table}

\begin{table}[h]
\begin{center}
\begin{tabular}{|c|c|}
\hline
Local critic network\\
\hline
Conv1, channel=512x512, kernel=3, stride=1, pad=1 \\
Spectral Normalization, LeakyRelu \\
Conv2, channel=512x128, kernel=2, stride=1, pad=1 \\
Spectral Normalization, LeakyRelu \\
Conv3, channel=128x128, kernel=2, stride=1, pad=1 \\
Spectral Normalization\\
\hline
\end{tabular}
\end{center}
\caption{The architecture of the local critic network}
\end{table}

\section*{Algorithm and Implementation Details}
The proposed Wasserstein distance based domain adaptation algorithm is summarized in Algorithm~\ref{alg:A}. We pre-train the VGG16 network on ImageNet, and fix the first three convolution layers, and then train the VGG16-based faster-RCNN on the source dataset. For all experiments, we resize the images to a shorter side of 600 and longer side to less than 1200. We also flip the images horizontally to augment the datasets. In Phase 1, we perform the global feature alignment as explained in Section 3.1 of main paper, and then fix the backbone feature generator VGG16. In Phase 2, we perform the local feature alignment by updating the shared region proposal network and classifier as explained in Section 3.2 of main paper. For both global and local alignment, we use an adam optimizer with a learning rate of $2 \times 10^{-4}$ and a $\beta$ of (0, 0.99). For the object detection loss, we also use adam optimizer but change the $\beta$ to (0.5, 0.99). We use a batch-size of $20$ for global alignment and $14$ for local alignment. 

\begin{algorithm}[H]
\caption{Wasserstein distance based domain adaptive Faster-RCNN}
\label{alg:A}
\begin{algorithmic}
\STATE {\textbf{Require}}: $\alpha$,$\gamma$ the learning rates. $c$, the clipping parameter. $n$, the batch size. $s_d$, the number of iterations to train discriminator.
\STATE {\textbf{Require}}: $\omega_g$, parameters for global critic, $\omega_l$, parameters for local critic, $\theta_s$, parameters for $M_s^g$, $\theta_t$, parameters for $M_t^g$, $\sigma$, parameters for $M^l$.

\STATE {\textbf{Phase 1}}:
\STATE {Pretrain the source Faster-RCNN on source data; Initialize $\theta_t$ with $\theta_s$}
\WHILE {$\theta_t$ has not converged}
\FOR {$t = 0,..., s_{d}$}
\STATE {Sample $\{X_s^i\}^n $ a batch from the source domain}
\STATE {Sample $\{X_t^i\}^n $ a batch from the target domain}
\STATE {Calculate the spectral normalization $\bar{\omega}_g$} of $\omega_g$
\STATE {$g_{\bar{\omega}_g} \leftarrow \nabla_{\bar{\omega}_g} [\frac{1}{n} \sum_{i} D^g(M_s^g(X_s^i)) - \frac{1}{n} \sum_{i} D^g(M_t^g(X_t^i))]$}

\STATE {$\omega_g \leftarrow \omega_g + \alpha \cdot Adam(\bar{\omega}_g, g_{\omega_g}) $}
\ENDFOR

\STATE {$g_{\theta_t} \leftarrow - \frac{1}{n} \sum_{i} D^g(M_t^g(X_t^i))$}
\STATE {$\theta_t \leftarrow \theta_t - \alpha \cdot Adam(\theta_t, g_{\theta_t}) $}

\ENDWHILE 

\STATE {\textbf{Phase 2}}
\WHILE {$M^l$ and $C$ has not converged}
\STATE {Sample $\{f_s^i\}^m $ a batch from the source domain}
\STATE {Sample $\{f_t^i\}^m $ a batch from the target domain}

\STATE {Calculate the spectral normalization $\bar{\omega_l}$ of $\omega_l$}

\STATE {$g_{\bar{\omega_l}} \leftarrow \nabla_{\bar{\omega_l}} [\frac{1}{nm} \sum_{i,j} D^l(M_{j}^{l}(f_s^i)) - \frac{1}{nm} \sum_{i,j} D^l(M_{j}^l(f_t^i))]$}
\STATE {$\omega_l\leftarrow \omega_l + \alpha \cdot Adam(\theta_l, g_{\bar{\omega_l}}) $}
\STATE {$g_{\sigma} \leftarrow \nabla_{\sigma} [\frac{1}{nm} \sum_{i,j} D^l(M_{j}^{l}(f_s^i)) - \frac{1}{nm} \sum_{i,j} D^l(M_{j}^l(f_t^i))$}
\STATE {$\sigma \leftarrow \sigma - \alpha \cdot Adam(\sigma, g_{\sigma}) + \gamma c \cdot clip(\nabla_{\sigma} \mathcal{L}_{det})$}

\ENDWHILE
\end{algorithmic}
\end{algorithm}

\end{document}